\documentclass[letterpaper, 10 pt, conference]{ieeeconf}  

\usepackage{graphicx}
\usepackage{cite}
\usepackage{amsmath}
\usepackage{amssymb}
\usepackage{booktabs} 
\usepackage{multirow}
\usepackage{caption}
\usepackage[table]{xcolor}
\usepackage{array}
\usepackage{tabularx}
\usepackage{makecell}
\usepackage{bigstrut}
\usepackage{algorithm}
\usepackage{algpseudocode}

\usepackage{xcolor}
\usepackage[skins,breakable]{tcolorbox}
\usepackage{caption}

\definecolor{ROOTReason}{HTML}{105CA1}
\definecolor{ROOTExecute}{HTML}{BC6200}
\definecolor{ROOTObserve}{HTML}{347719}

\newtcolorbox{ROOTpromptbox}[2]{%
  enhanced jigsaw,
  breakable,
  width=\linewidth,
  colback=white,
  colframe=#1!65!black,
  colbacktitle=#1!9!white,
  coltitle=#1!65!black,
  title={#2},
  title after break={#2\enspace(continued)},
  fonttitle=\small\bfseries,
  fontupper=\small,
  boxrule=0.45pt,
  arc=0.7mm,
  left=5pt,
  right=5pt,
  top=4pt,
  bottom=4pt,
  toptitle=3pt,
  bottomtitle=3pt,
  before skip=8pt,
  after skip=0pt,
  lines before break=6,
  before upper={%
    \setlength{\parindent}{0pt}%
    \setlength{\parskip}{3pt}%
    \interlinepenalty=10000
    \raggedright
  }
}

\newcommand{\ROOTpromptfield}[1]{%
  \par\smallskip
  \noindent\textbf{#1}\par\nobreak
}

\makeatletter
\newcommand{\fs@ROOTruled}{%
    \fs@ruled
    \let\ROOToriginalpre\@fs@pre
    \def\@fs@pre{\vspace{0.5em}\ROOToriginalpre}%
}
\makeatother

\floatstyle{ROOTruled}
\restylefloat{algorithm}
\floatstyle{ruled}

\definecolor{Ransalu}{RGB}{228,26,28}
\definecolor{Xin}{RGB}{0,0,255}
\definecolor{Eren}{RGB}{77,175,74}
\definecolor{Aditya}{RGB}{152,78,163}

\IEEEoverridecommandlockouts          

\title{\textsc{ROOT}: Discovering Rewards for User-Specified Embodied Behaviors
}

\author{Eren Sadikoglu$^{1}$, Aditya Taparia$^{2}$, Xinyuan Liu$^{2}$, and Ransalu Senanayake$^{2}$%
\thanks{$^{1}$Robotics and Autonomous Systems, Arizona State University, Tempe, Arizona, USA}%
\thanks{$^{2}$School of Computing and Augmented Intelligence, Arizona State University, Tempe, Arizona, USA {\tt\small \{esadikog, ataparia, xinyua11, ransalu\}@asu.edu}}%
}

\begin{document}

\maketitle
\thispagestyle{empty}
\pagestyle{empty}

\begin{abstract}

Reinforcement learning for embodied control remains constrained by the difficulty of reward specification. Although recent large language model (LLM)-based methods can synthesize reward functions from natural-language descriptions, they often fail to capture subtle behavioral properties that humans care about, such as natural gait, posture, and movement style. This limitation arises because many desired behaviors are easier to recognize visually than to encode in a reward function. We introduce Reward Optimization via Observable Trees (ROOT), a framework for discovering reward functions that align learned policies with user-specified embodied behaviors. Rather than relying solely on scalar training statistics, ROOT casts reward design as an observation-guided search over a persistent experiment tree that stores reward programs, trained policies, and rollout observations, together with behavioral insights distilled by a video-language model, to diagnose behavioral failures and guide subsequent reward refinements. We evaluate ROOT on \textbf{seven} tasks across four embodiments: simulated Hopper, HalfCheetah, Ant, Unitree Go2, and as well as the real-world Unitree Go2. ROOT produces behaviors that better align with user intent than those generated by existing LLM-based reward-generation methods, achieving up to \textbf{86.8\%} locomotion-completeness accuracy and improving Vid-LLM behavioral alignment from \textbf{3.56/5} to \textbf{4.14/5}, a \textbf{16.5\%} improvement over baselines. Human evaluations further support these results, with ROOT preferred in \textbf{51--63\%} of pairwise comparisons.
\end{abstract}

\section{Introduction}
\label{sec:intro}

Deep reinforcement learning (RL) has enabled increasingly capable robotic behaviors in recent years~\cite{luo2020carl, han2024lifelike, radosavovic2024real, zhang2026falcon}. Learning these behaviors depends on reward functions that translate the desired task into a training objective. This translation is relatively straightforward when the objective is well-defined and directly measurable, such as reaching a target position or maintaining a desired speed. However, reward design becomes more challenging when the desired behavior also involves qualitative properties, such as posture, limb coordination, and movement style. For example, asking a quadruped to trot forward requires both forward motion and coordinated leg movements in a recognizable trotting motion. Such properties are often easier to recognize visually than to specify through hand-crafted reward terms.

\begin{figure}[t]
    \centering
    \includegraphics[width=0.9\linewidth]{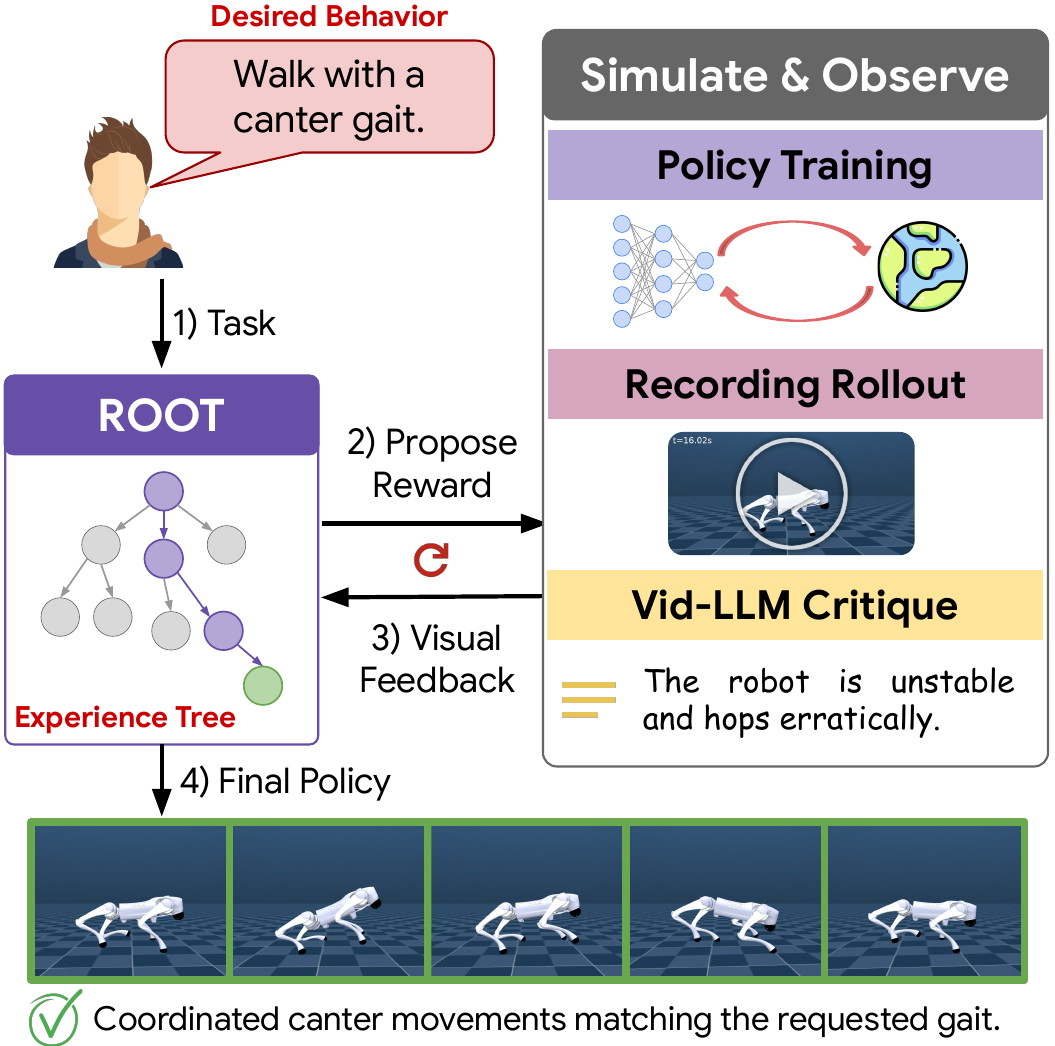}
    \caption{{Overview of ROOT.} Given a user-specified behavior, ROOT explores a tree of candidate reward programs. Each candidate is visually audited and saved. The learned tree guides later reward generation toward policies that better match the intended behavior.}
    \label{fig:intro}
    \vspace{-1em}
\end{figure}

To reduce this burden of manual reward engineering, recent methods such as Eureka~\cite{ma2024eureka} and Text2Reward~\cite{xie2024text2reward} use large language models (LLMs) to generate executable reward programs from natural-language instructions and environment descriptions.  These approaches automate reward design across diverse control tasks. However, generated rewards may not fully
capture the requested behavioral properties, allowing policies to earn high returns while deviating from the user's intent. A quadruped instructed to trot forward, for instance, may learn to propel itself using its front legs while dragging its hind legs, earning a high reward for forward motion without exhibiting the alternating diagonal leg coordination that defines a trot. We refer to this mismatch between the user's intended behavior and the learned policy's behavior as the \emph{behavioral gap}. Because scalar training statistics alone may not reveal this gap, recent methods incorporate visual feedback to guide reward refinement~\cite{chen_elemental_2025,cuzin2025viral}. Yet identifying a behavioral failure is only part of the refinement process, this feedback must also inform which past reward to revisit and what modification to test next.

To that end, we present Reward Optimization via Observable Trees (\textsc{ROOT}), which organizes reward discovery as an observation-guided search over a persistent tree of experience (Fig.~\ref{fig:intro}). Each experience connects a reward program and its trained policy with rollout observations, scalar diagnostics, and distilled behavioral insights, while branches represent alternative reward refinements. A multi-agent reasoning framework uses this accumulated evidence to diagnose failures, revisit promising rewards, and propose targeted refinements that address the behavioral gap. Across seven tasks on four simulated embodiments (Hopper, HalfCheetah, Ant, and Unitree Go2), with additional validation on a physical Go2, \textsc{ROOT} produces policies that better align with user intent and are preferred by human evaluators over existing LLM-based reward-design baselines. Our contributions are as follows:

\begin{enumerate}
    \item We formulate reward discovery as a search over an observable ``experience tree'' that stores reward programs, trained policies, rollout observations, scalar diagnostics, hypotheses, and behavioral insights, allowing subsequent generation to build on past experience.
    \item We introduce \textsc{ROOT}, a multi-agent framework which use past evidence from the experience tree to diagnose failures, revisit promising rewards, and propose targeted refinements.
    \item We demonstrate that \textsc{ROOT} improves alignment with user intent over existing reward-design baselines across multiple embodied tasks, and show how past experience contributes to these improvements.
\end{enumerate}
\section{Related Work}

\label{sec:related}
\paragraph{LLM-guided reward design} Reward design remains a major bottleneck in embodied RL, with 92\% of practitioners relying on manual reward tuning and 89\% reporting sub-optimal rewards~\cite{booth2023perils}. LLMs facilitate reward design by translating task descriptions into reward signals~\cite{kwon2023reward}, structured specifications~\cite{yu2023language}, or executable code~\cite{xie2024text2reward}. EUREKA~\cite{ma2024eureka} refines sampled reward programs using training statistics and optional human feedback, while REvolve~\cite{hazra2025revolve} and CARD~\cite{sun2024largelanguagemodeldrivenreward} explore evolutionary refinement and trajectory-preference evaluation. Auto MC-Reward~\cite{li_auto_2024} separates reward generation, code critique, and trajectory analysis to guide revisions. More structured approaches include R*~\cite{li_r_2025}, which evolves modular reward structures and learns their parameters from trajectory preferences, and ORSO~\cite{zhang_orso_2025}, which interleaves reward selection with policy optimization. These approaches automate reward synthesis and refinement, motivating further study of how experimental feedback should guide subsequent reward-design decisions.

\paragraph{Structured search and reuse of experimental history} 
Reflexion~\cite{shinn_reflexion_2023} retains textual reflections in episodic memory to guide subsequent trials. In reward design, RF-Agent~\cite{gao_rf-agent_2025} formulates reward optimization as Monte Carlo tree search, storing reward programs and training feedback in tree nodes and using scalar task evaluations, self-verification, and tree statistics to guide selection and expansion. ROOT emphasizes behavioral evidence in this process: each experiment links a refinement plan to the resulting policy behavior, while parent--child
relationships preserve the reference reward. Its reasoning procedure uses these records to jointly select an earlier reward to revisit and a modification to test, informed by alternative refinements and observed
regressions across branches. We investigate how access to accumulated experimental
evidence and visual feedback contributes to reward discovery.

\paragraph{Visual feedback in reward design} A separate line of work uses vision language models (VLMs) to provide behavioral supervision through several mechanisms. VLM-RMs~\cite{rocamonde2024vision} derive rewards from image--text alignment, RL-VLM-F~\cite{wang2024rl} learns reward models from visual preferences, and VLM-CaR~\cite{venuto2024code} generates executable rewards to reduce repeated VLM inference. Related preference-based approaches include LAPP~\cite{jian_lapp_2025}, which learns from LLM preferences over state--action trajectories, and ReWiND~\cite{zhang_rewind_2025}, which learns language-conditioned rewards from demonstrations and adapts policies without additional task demonstrations. Closest to ROOT, ELEMENTAL~\cite{chen_elemental_2025} combines visual demonstrations, generated reward features, inverse reinforcement learning, and iterative self-reflection. VIRAL~\cite{cuzin2025viral} refines reward code using training statistics and optional human or video-language-model feedback. ROOT uses visual critique of its own trained-policy rollouts as experimental evidence, without requiring task demonstrations. These assessments remain linked to the corresponding reward refinements and inform subsequent search decisions.


\begin{figure*}[t]
    \centering
    \vspace{0.5em}
    \includegraphics[width=\linewidth]
        {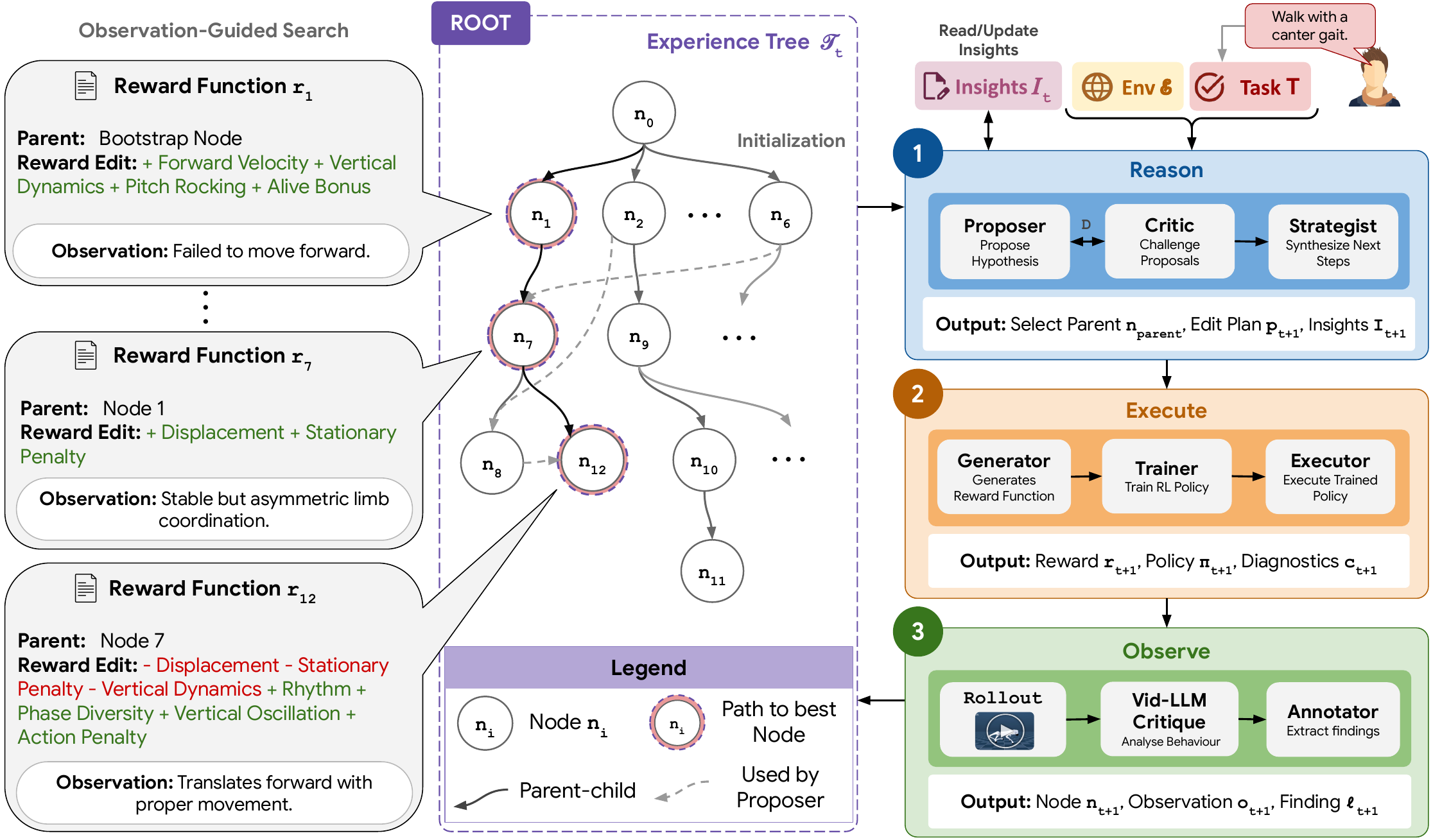}
    \caption{
        {Observation-guided reward search with ROOT.} Each tree expansion selects a reference reward and a refinement hypothesis, generates a reward program, trains a policy, and         records the observed outcome. The experience tree preserves both partial successes and unsuccessful refinements, allowing evidence from earlier experiments and other branches to condition subsequent proposals. A Vid-LLM supplies visual feedback during search.
    }
    \label{fig:method}
    \vspace{-1em}
\end{figure*}

\section{Methodology}
\label{sec:method}

Given an environment context $\mathcal{E}$ and a task $T$, \textsc{ROOT} searches for a reward program that induces a policy aligned with the user's intended task. For an executable reward $r$ and a fixed policy-training procedure $\mathsf{Train}$, we formalize the behavioral gap introduced in Sec.~\ref{sec:intro} as,
\begin{equation}
\begin{aligned}
    \pi_r
        &= \mathsf{Train}(\mathcal{E},r),\\
    \Delta_H(r;T)
        &= \mathbb{E}_{\tau\sim P_{\mathcal{E}}^{\pi_r}}
           \left[d_H(\tau,T)\right],
\end{aligned}
\label{eq:root:gap}
\end{equation}
where $\pi_r$ is the policy obtained from a training run, $P_{\mathcal{E}}^{\pi_r}$ is its rollout distribution, and $d_H(\tau,T)\geq 0$ denotes the discrepancy between a rollout and the user's intended behavior. The goal is to identify a reward with a small $\Delta_H$ within a search budget $B$.

The discrepancy $d_H$ is an implicit human criterion, so $\Delta_H$ is not directly evaluated during search. Instead, \textsc{ROOT} uses a Vid-LLM to identify behavioral discrepancies in policy rollouts. These assessments guide reward refinements intended to reduce the behavioral gap, while an experience tree links the proposed changes to their observed outcomes. Evidence of improvements and regressions then informs which reward design to revisit and what modification to test next. The assessments and experiment history thus provide qualitative guidance toward the objective in Eq.~\eqref{eq:root:gap}. Figure~\ref{fig:method} illustrates this process.

\subsection{The Experience Tree}
\label{sec:method:tree}

At iteration $t$, the search state comprises an experience tree $\mathcal{T}_t=(\mathcal{V}_t,\mathcal{A}_t)$ and an auxiliary insights document $I_t$. Each successfully evaluated non-root node stores:
\begin{equation}
    n_i=(r_i,\pi_i,p_i,c_i,o_i,\ell_i),
    \label{eq:root:node}
\end{equation}
where $r_i$ is the reward program used for training, $\pi_i$ is the saved policy, and $p_i$ is a \emph{refinement plan} specifying the proposed reward changes and their expected behavioral effect. The diagnostic record $c_i$ contains training statistics and quantitative evaluation metrics, $o_i$ contains rollout observations, and $\ell_i$ summarizes the experimental findings and their implications for subsequent refinement. Failed attempts are also retained, with failure diagnostics saved in $c_i$.

An edge $(n_j,n_i)$ records that $n_j$ was selected as the reference for the refinement tested at $n_i$. Parent-child records therefore associate a proposed reward change with its observed outcome, while siblings retain alternative refinements of the same reference. Earlier nodes remain available after their descendants have been evaluated, allowing search to return to a partial success when subsequent changes regress.

\paragraph{Initialization} \textsc{ROOT} first expands the task $T$ into an environment-grounded specification $\bar{T}$ that retains the intended task while identifying relevant physical signals and anticipated failure modes. It initializes $I$ from this specification and the environment context, and creates a root node $n_0$. The first $N_{\mathrm{init}}$ experiments generate bootstrap reward candidates using varied initial design guidance within $\mathsf{Generate}$. These candidates have no parent reward or refinement plan, so $n_{\mathrm{parent}}=n_0$ and $p_i=\varnothing$. Their experiment records form the initial children of the synthetic root and provide evidence for subsequent refinement.

\paragraph{Experiment Memory and Expansion} Retrieval tools allow the reasoning module to inspect relevant experiment records and compare findings across branches without loading the full history into context. The tree preserves experiment-specific evidence, while $I_t$ maintains design guidance distilled across experiments and revised during reasoning. Together, these records inform both the selection of a node to expand and the changes for its refinement.

\subsection{Observation-Guided Search}
\label{sec:method_loop}

Each refinement iteration comprises three stages: \textit{Reason} selects a reference and a refinement plan, \textit{Execute} tests the proposed reward through policy training, and \textit{Observe} records its rollout behavior. The budget $B$ counts experiment attempts, including bootstrap and failed attempts.

\paragraph{Reason} Conditioned on $\bar{T}$ and $\mathcal{E}$, the reasoning module uses the relationships encoded in $\mathcal{T}_t$ to guide the next expansion. Parent-child comparisons connect reward refinements to their observed behavioral effects, while sibling comparisons reveal the outcomes of alternative refinements from a shared reference. Evidence across branches further allows the search to revisit earlier partial successes when subsequent refinements regress. Drawing on these comparisons, the module selects a parent $n_{\mathrm{parent}}$, produces a refinement plan $p_{t+1}$, and updates the insights document to $I_{t+1}$. The plan identifies which reward-design elements to preserve and which remaining behavioral discrepancies to target.

We implement this stage with three LLM roles. A \emph{Proposer} examines the experimental evidence and suggests refinements. A \emph{Critic} challenges the proposal by checking its reasoning against prior attempts and overlooked partial successes. They alternate for up to $D$ rounds, after which a \emph{Strategist} synthesizes the discussion into the parent selection, refinement plan, and updated insights.

\paragraph{Execute} The generator receives $\bar{T}$, $\mathcal{E}$, $p_{t+1}$, and the selected parent's experiment record, including its reward implementation. The environment context exposes the observation and action spaces, joint names, and body structure. The generator translates this information into a candidate reward program.

The execution procedure $\mathsf{Execute}$ validates the program and applies bounded repairs before training a fresh PPO policy using $\mathsf{Train}$. On success, it returns the reward implementation actually used for training, the resulting policy, and the diagnostic record. Reward-component curves are recorded during training, followed by quantitative policy evaluation. Since every policy is trained from scratch, tree edges represent reward-design lineage. If execution fails, the diagnostic record captures the failure and $\pi_{t+1}=\varnothing$.

\paragraph{Observe} Following successful execution, the policy is rendered from multiple views. The Vid-LLM assesses these rollout videos against $\bar{T}$ to produce $o_{t+1}$, describing the achieved behavior and remaining discrepancies. A lightweight annotator combines these observations with the diagnostics, parent findings, and current insights to produce $\ell_{t+1}$. This record summarizes whether the expected effect was observed and what the outcome suggests for further refinement.

The resulting experiment record is added as a child of $n_{\mathrm{parent}}$, including when its outcome is a regression or an execution failure. These records remain available to inform subsequent parent selection and reward generation. After the budget is exhausted, $\mathsf{Select}_{\mathrm{align}}$ compares the task-alignment descriptions recorded for successfully evaluated policies and selects the node whose policy is judged to best match the requested behavior $T$. The corresponding policy $\hat{\pi}$ is then returned. Algorithm~\ref{alg:root} summarizes the procedure.

\begin{algorithm}[t]
\caption{\textsc{ROOT}: Observation-Guided Reward Search}
\label{alg:root}
\begin{algorithmic}[1]
\Require Task $T$, environment $\mathcal{E}$,
         bootstrap count $N_{\mathrm{init}}$,
         experiment budget $B\geq N_{\mathrm{init}}\geq 1$
\Ensure Selected policy $\hat{\pi}$, or failure

\State $\bar{T}\gets\mathsf{Expand}(T,\mathcal{E})$
\State $I\gets\mathsf{InitializeInsights}(\bar{T},\mathcal{E})$
\State Initialize tree $\mathcal{T}$ with synthetic root $n_0$
\State Set $r_0=\ell_0=\varnothing$

\For{$i=1,\ldots,B$}
    \If{$i\leq N_{\mathrm{init}}$}
        \Comment{Bootstrap}
        \State $(n_{\mathrm{parent}},p_i)
            \gets(n_0,\varnothing)$
    \Else
        \State $(n_{\mathrm{parent}},p_i,I)\gets$
            $\mathsf{Reason}(\mathcal{T},I;\bar{T},\mathcal{E})$
    \EndIf

    \State $r_i\gets
        \mathsf{Generate}
        (\bar{T},\mathcal{E},p_i,n_{\mathrm{parent}})$
    \State $(r_i,\pi_i,c_i)\gets
        \mathsf{Execute}(\mathcal{E},r_i)$

    \State $o_i\gets\varnothing$
    \If{$\pi_i\neq\varnothing$}
        \State $v_i\gets\mathsf{Render}(\pi_i)$
        \State $o_i\gets
            \mathsf{Observe}(v_i,\bar{T},p_i,c_i)$
    \EndIf

    \State $\ell_i\gets$
        $\mathsf{Annotate}
        (o_i,c_i,p_i,\ell_{\mathrm{parent}},I)$

    \State Add $n_i=(r_i,\pi_i,p_i,c_i,o_i,\ell_i)$
    \Statex \hspace{\algorithmicindent}
        as a child of $n_{\mathrm{parent}}$ in $\mathcal{T}$
\EndFor

\State $\hat{n}\gets
    \mathsf{Select}_{\mathrm{align}}(\mathcal{T},T)$
\State \Return policy $\hat{\pi}$ stored at $\hat{n}$
\end{algorithmic}
\end{algorithm}
\begin{table*}[t]
\centering
\vspace{0.5em}
\caption{Behavioral Alignment across Simulated Embodiments. We report task performance and Vid-LLM alignment; higher is better. Best and second-best scores are bold and underlined. Locomotion completeness accuracy uses task-specific metrics defined in Sec.~\ref{sec:exp-setup} and reports mean and standard deviation across three evaluation seeds. Alignment reports mean Vid-LLM scores. $^\ast$native score denotes mean velocity multiplied by mean survival. \textbf{ROOT achieves the best behavioral alignment while maintaining comparable or higher task performance across tasks.}}
\label{tab:behavioral_alignment}

\begingroup

\setlength{\tabcolsep}{2pt}
\renewcommand{\arraystretch}{1.1}

\setlength{\bigstrutjot}{3pt}

\def\ROOTscore#1#2{%
    \ensuremath{#1\!\pm\!#2}%
}
\def\ROOTbestscore#1#2{%
    \ensuremath{\mathbf{#1\!\pm\!#2}}%
}
\def\ROOTsecondscore#1#2{%
    \ensuremath{\underline{#1\!\pm\!#2}}%
}

\def\ROOTrobotlabel#1#2{%
    \multirow{-#1}{*}{%
        \rotatebox[origin=c]{90}{\textit{#2}}%
    }%
}

\def\tasklabel#1{%
    \multirow{-2}{=}{\raggedright #1}%
}

\def\ROOTtallrobot#1{%
    \multirow{-2}[tb2]{*}{%
        \rotatebox[origin=c]{90}{\textit{#1}}%
    }%
}

\def\ROOTtalltask#1{%
    \multirow{-2}[tb2]{=}{\raggedright #1}%
}

\arrayrulecolor{black}

\begin{tabularx}{\textwidth}{
    >{\centering\arraybackslash}p{0.95cm}
    >{\centering\arraybackslash}p{4.1cm}
    >{\centering\arraybackslash}p{1.7cm}
    *{5}{>{\centering\arraybackslash}X}
    |
    >{\centering\arraybackslash}X
}
\toprule
\multicolumn{1}{c}{\multirow{2}{*}{\textbf{Robot}}}
& \multirow{2}{*}{\textbf{Task}}
& \multirow{2}{*}{\textbf{Metric}}
& \multirow{2}{*}{\textbf{Eureka}}
& \textbf{Text2}
& \textbf{Language}
& \multirow{2}{*}{\textbf{CARD}}
& \multirow{2}{*}{\textbf{LAPP}}
& \textbf{ROOT} \\
& & & & \textbf{Reward} & \textbf{2Reward}
& & & \textbf{(Ours)} \\
\midrule


& & Align. Score
& \textbf{4.67} & 4.50 & \underline{4.61} & \textbf{4.67} & 3.94 & \textbf{4.67} \\

& \tasklabel{Walk forward with a \textbf{trot} gait.}
& Acc. (\%)
& \ROOTsecondscore{79.6}{0.7} & \ROOTscore{78.1}{0.1} & \ROOTscore{9.8}{0.3} & \ROOTscore{79.5}{0.4} & \ROOTscore{0.0}{0.0} & \ROOTbestscore{86.8}{0.3} \\

\arrayrulecolor{black!20}
\cline{2-9}
\arrayrulecolor{black}

& & Align. Score
& \textbf{4.56} & 4.39 & 4.39 & \underline{4.50} & 4.17 & \underline{4.50} \\

& \tasklabel{Walk forward with a \textbf{pace} gait.}
& Acc. (\%)
& \ROOTscore{74.9}{0.2} & \ROOTsecondscore{77.5}{0.3} & \ROOTscore{13.7}{1.5} & \ROOTscore{76.1}{0.5} & \ROOTscore{7.48}{0.9} & \ROOTbestscore{78.4}{0.1} \\

\arrayrulecolor{black!20}
\cline{2-9}
\arrayrulecolor{black}

& & Align. Score
& 2.78
& 2.72
& 2.22
& \underline{3.50}
& 3.22
& \textbf{3.61} \\

\ROOTrobotlabel{7}{Unitree Go2}
& \tasklabel{Move forward with a \textbf{canter} gait.}
& Acc. (\%)
& \ROOTscore{0.8}{0.2}
& \ROOTscore{67.8}{2.4}
& \ROOTscore{1.4}{1.1}
& \ROOTscore{19.1}{1.4}
& \ROOTbestscore{70.8}{3.6}
& \ROOTsecondscore{70.5}{4.0} \\
\midrule


& & Align. Score
& 2.56
& 2.72
& \underline{3.67}
& 2.56
& 3.39
& \textbf{4.44} \\

& \tasklabel{Move sideways with a \textbf{crab-crawling} gait.}
& Acc. (\%)
& \ROOTscore{1.8}{0.1}
& \ROOTbestscore{65.2}{13.5}
& \ROOTscore{32.3}{1.6}
& \ROOTscore{1.4}{0.5}
& \ROOTscore{31.2}{22.7}
& \ROOTsecondscore{47.67}{2.64} \\

\arrayrulecolor{black!20}
\cline{2-9}
\arrayrulecolor{black}

& & Align. Score
& 2.06
& \underline{3.56}
& \underline{3.56}
& 1.17
& \textbf{3.89}
& \textbf{3.89} \\

\ROOTrobotlabel{4}{Ant}
& \tasklabel{Move forward with a \textbf{spider-creeping} gait.}
& Acc. (\%)
& \ROOTscore{1.0}{0.8}
& \ROOTscore{15.9}{9.8}
& \ROOTscore{35.0}{2.5}
& \ROOTscore{12.6}{14.6}
& \ROOTbestscore{54.4}{2.8}
& \ROOTsecondscore{35.50}{4.8} \\
\midrule


& & Align. Score\bigstrut[t]
& 1.39
& 1.33
& \underline{2.11}
& \textbf{3.56}
& 1.39
& \textbf{3.56} \\

\ROOTtallrobot{Hopper}
& \ROOTtalltask{\textbf{Hop forward} as fast as possible.}
& Native Score$^\ast$\bigstrut[b]
& \ROOTscore{0.19}{0.04}
& \ROOTscore{0.73}{0.00}
& \ROOTscore{0.75}{0.08}
& \ROOTsecondscore{1.71}{0.01}
& \ROOTscore{0.10}{0.00}
& \ROOTbestscore{1.82}{0.45} \\
\midrule


& & Align. Score\bigstrut[t]
& \underline{4.11}
& \underline{4.11}
& \textbf{4.33}
& \textbf{4.33}
& \textbf{4.33}
& \textbf{4.33} \\

\ROOTtallrobot{\shortstack{Half\\Cheetah}}
& \ROOTtalltask{\textbf{Run forward} as fast as possible.}
& Native Score$^\ast$\bigstrut[b]
& \ROOTscore{1.79}{0.54}
& \ROOTbestscore{7.08}{0.00}
& \ROOTscore{6.50}{0.01}
& \ROOTscore{6.75}{0.07}
& \ROOTscore{3.42}{0.03}
& \ROOTsecondscore{6.99}{0.04} \\

\bottomrule
\end{tabularx}
\endgroup
\end{table*}

\section{Experiments}
\label{sec:experiment}

We evaluate \textsc{ROOT} in terms of behavioral alignment, task performance, and reward-search cost, and assess whether the learned behaviors transfer to a physical robot. Our experiments address four research questions:

\begin{description}
    \setlength{\itemsep}{2pt}
    \setlength{\parsep}{0pt}

    \item[\textbf{RQ1:}] Can \textsc{ROOT} improve behavioral alignment while maintaining or improving task performance?

    \item[\textbf{RQ2:}] How do the experience tree and VLM-based behavioral feedback contribute to improved alignment?

    \item[\textbf{RQ3:}] How does \textsc{ROOT} compare with the baselines in behavioral alignment for a given search cost?

    \item[\textbf{RQ4:}] How well do policies trained for behavioral alignment in simulation retain the requested behaviors when deployed on a physical robot?
\end{description}

\subsection{Experimental Setup}
\label{sec:exp-setup}

\paragraph{Tasks and environments}

\begin{figure*}[t]
\vspace{0.5em}
\centering
\includegraphics[width=\linewidth]{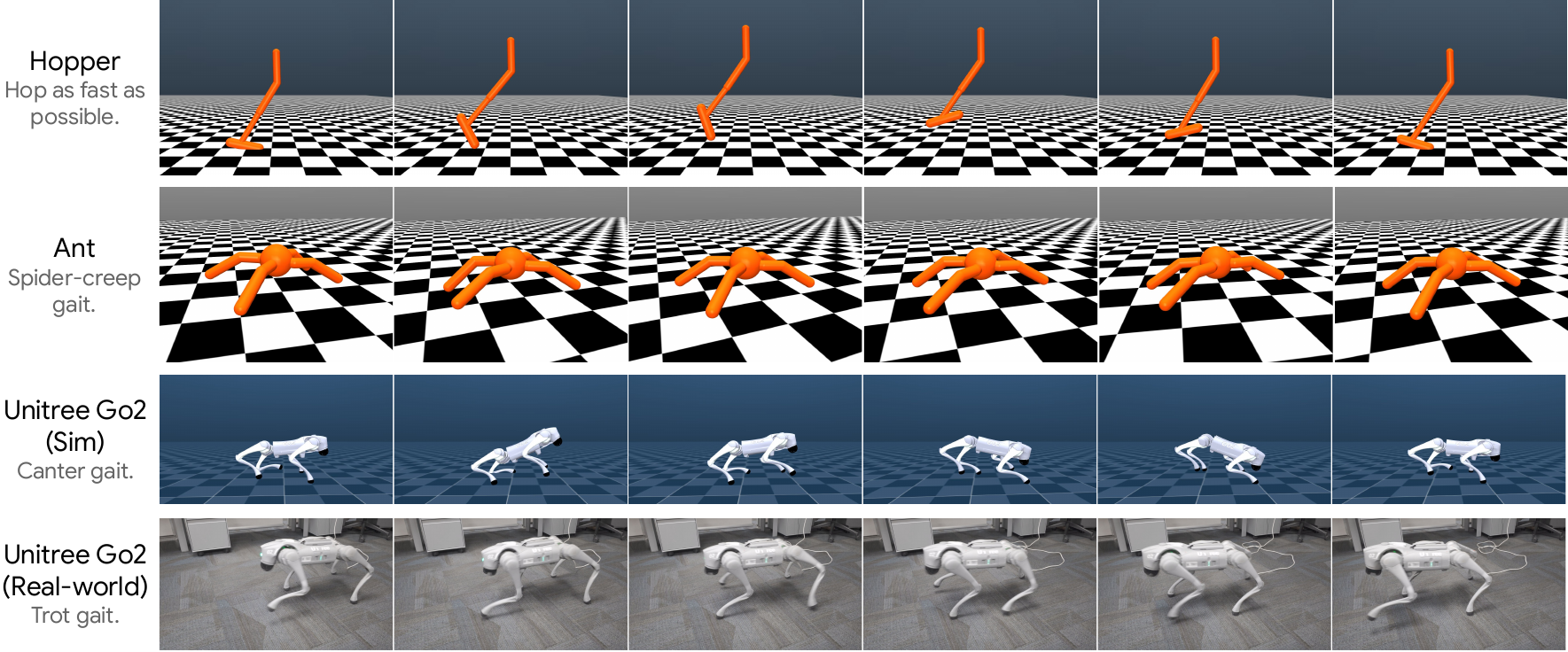}
\caption{{Qualitative rollouts of policies learned with \textsc{ROOT}.} From top to bottom: Hopper hopping, Ant spider-creeping, and Unitree Go2 cantering in simulation, followed by a simulation-trained trotting policy deployed on a physical Unitree Go2. Frames progress from left to right.}
\label{fig:rollouts}
\end{figure*}

As shown in Fig.~\ref{fig:rollouts}, we evaluate \textsc{ROOT} in two distinct simulators and validate the resulting policy on a physical Unitree Go2 quadruped. The simulation evaluation covers seven simulated tasks across four embodiments: Ant crab-crawl and spider-creep; Unitree Go2 pace, trot, and canter; and forward running with Hopper and HalfCheetah. Hopper and HalfCheetah provide conventional locomotion benchmarks for comparison with prior reward-design methods. The remaining tasks require specific direction, posture, or limb-coordination properties beyond forward progress, allowing us to examine alignment with qualitatively specified behaviors. To emphasize the challenge of the ``behavioral gap,'' all tasks are specified by natural language rather than formal mathematical objectives, summarized in Table~\ref{tab:behavioral_alignment}; quantitative evaluation criteria are defined separately below. Hopper, HalfCheetah, both Ant tasks, and Go2 canter use MJX (MuJoCo XLA)~\cite{todorov2012mujoco}, while Go2 pace and trot use Isaac Lab~\cite{mittal2025isaac}. We additionally deploy a learned trot policy on a physical Unitree Go2 to assess sim-to-real transfer.

\paragraph{Baselines and implementation}
We compare \textsc{ROOT} against five state-of-the-art LLM-based reward engineering frameworks: Eureka~\cite{ma2024eureka}, Text2Reward~\cite{xie2024text2reward}, L2R~\cite{yu2023language}, CARD~\cite{sun2024largelanguagemodeldrivenreward}, and LAPP~\cite{jian_lapp_2025}. Within each simulator backend, methods share task descriptions, PPO configurations, and per-candidate training budgets: 20M environment steps on MJX (30M for Go2 canter) and approximately 98M on Isaac Lab. Candidate counts and policy-selection rules follow method-specific protocols. We evaluate a preference-only adaptation of LAPP on both simulator backends, removing the handcrafted environment reward ($r=\beta r_p$) and matching its environment-step budget to \textsc{ROOT}’s total training budget. All methods use Claude 4.5 Sonnet as the same backbone language model for reasoning and reward code generation. In \textsc{ROOT}’s observation stage, we employ Gemini 3.1 Pro as the Vid-LLM model to evaluate rendered rollout videos and produce structured behavioral feedback.

\paragraph{Evaluation Metrics}
\label{sec:exp-eval-metrics}
We compute task metrics from simulator states independently of the generated reward. For Hopper and HalfCheetah, we report \emph{native score}: mean forward velocity multiplied by mean survival. For gait tasks, we report
\emph{locomotion completeness accuracy}, defined as contact-based
accuracy during locomotion for Ant; the product of survival,
velocity tracking, contact coordination, and smoothness for
Go2 pace/trot; and the fraction of detected cycles matching
the target footfall pattern, weighted by survival. These tasks use a $0.5\,\mathrm{m/s}$ forward command with pushes disabled. Metrics are compared within tasks.

We report mean and standard deviation across three evaluation seeds using a $1{,}000$-step horizon. This uncertainty reflects rollout variability for the selected policy. The \emph{alignment score} is the mean Vid-LLM rating across
Seed 2.1 Turbo, Gemma 4 31B, and GLM-4.6V, which assess posture/style and limb coordination on a 1--5 scale using three-view videos, task descriptions, and simulator-derived
facts. 

\subsection{RQ1: Behavioral Alignment and Task Performance}
\label{sec:rq1}
\begin{figure}[t]
    \centering
    \includegraphics[width=\linewidth]{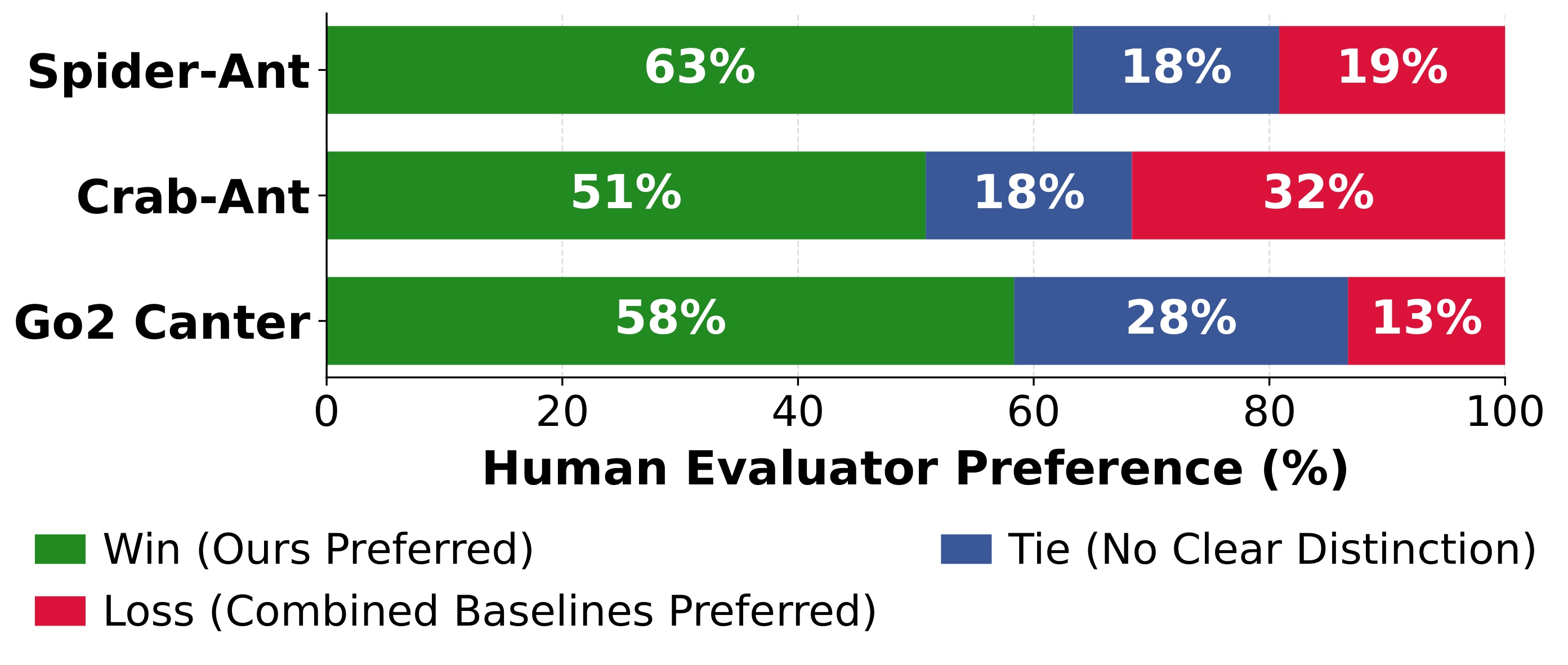}
    \caption{Human preferences across pairwise comparisons of \textsc{ROOT}
against pooled baselines on three behaviors.}
    \label{fig:human_preference}
    \vspace{-1em}
\end{figure}

Table~\ref{tab:behavioral_alignment} shows that \textsc{ROOT} remains competitive across the task suite, ranking first or second in mean task performance on all seven tasks. The strongest baseline varies across behaviors, whereas \textsc{ROOT} performs well on both forward locomotion and specified gaits. Human evaluation provides a focused assessment of crab-crawl, spider-creep, and canter (Fig.~\ref{fig:human_preference}),
favoring \textsc{ROOT} over the pooled baselines on all three. Notably, this preference holds on the Ant tasks despite \textsc{ROOT}'s lower contact-based accuracy than the leading methods,
suggesting that contact agreement alone does not fully capture perceived adherence to the requested behavior.

Across the full suite, the VLM panel assigns \textsc{ROOT} the highest or joint-highest alignment score on six of seven tasks, averaging 4.14 versus 3.56 for the strongest baseline. However, favorable ratings for LAPP on pace and trot despite low locomotion completeness accuracy show that VLM scores alone cannot establish correct coordination. We therefore interpret these scores alongside state-based metrics and
human preferences: together, they support competitive task performance and favorable perceived alignment.

\begin{table}[t]
\centering
\caption{We compare \textsc{ROOT} with variants that remove experimental history (or tree) and both history and Vid-LLM feedback during reward search. We report locomotion completeness accuracy (mean $\pm$ SD across three evaluation seeds)
and mean Vid-LLM alignment.) and mean Vid-LLM alignment; higher is better. \textsc{ROOT} results are bold.}
\label{tab:ablations}

\begingroup
\setlength{\tabcolsep}{4pt}
\renewcommand{\arraystretch}{1.1}

\def\ROOTboldscore#1#2{%
    \ensuremath{\mathbf{#1\!\pm\!#2}}%
}

\resizebox{\columnwidth}{!}{%
\begin{tabular}{llccc}
\toprule
\multirow{2}{*}{\textbf{Configuration}}
& \multirow{2}{*}{\textbf{Metric}}
& \multicolumn{2}{c}{\textbf{Ant}}
& \textbf{Unitree Go2} \\
\cmidrule(lr){3-4}
\cmidrule(lr){5-5}
& & \textbf{Crab} & \textbf{Spider} & \textbf{Canter} \\
\midrule

\multirow{2}{*}{\shortstack[l]{\textbf{\textsc{ROOT}}\\ \textbf{(Full)}}}
& \textbf{Acc. (\%)}
& \ROOTboldscore{47.67}{2.64}
& \ROOTboldscore{35.50}{4.8}
& \ROOTboldscore{70.5}{4.0} \\
& \textbf{Align. Score}
& \textbf{4.44}
& \textbf{3.89}
& \textbf{3.61} \\
\midrule

\multirow{2}{*}{\shortstack[l]{No History\\(No Tree)}}
& Acc. (\%)
& $2.97 \pm 2.4$
& $9.63 \pm 11.1$
& $19.63 \pm 0.35$ \\
& Align. Score & 2.83 & 2.67 & 2.89 \\
\midrule

\multirow{2}{*}{\shortstack[l]{No Vid-LLM\\and History}}
& Acc. (\%)
& $0.97 \pm 0.06$
& $3.40 \pm 2.97$
&  $1.37 \pm 0.06$ \\
& Align. Score & 2.56 & 1.89 & 1.06 \\

\bottomrule
\end{tabular}%
}
\endgroup
\end{table}

\begin{figure*}[t]
    \centering
    \begin{minipage}[t]{0.48\textwidth}
        \centering
        \includegraphics[width=\linewidth]{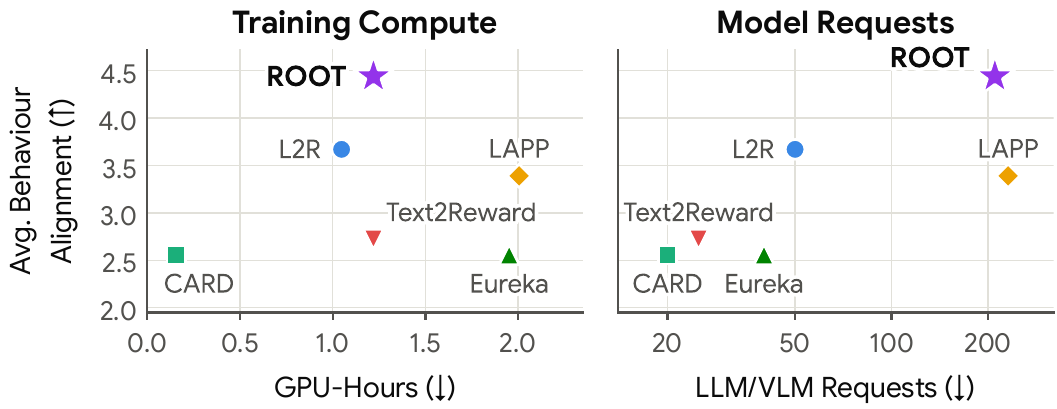}
        \par\smallskip
        {(a) Ant: Crab-crawling}
    \end{minipage}
    \hfill
    \begin{minipage}[t]{0.48\textwidth}
        \centering
        \includegraphics[width=\linewidth]{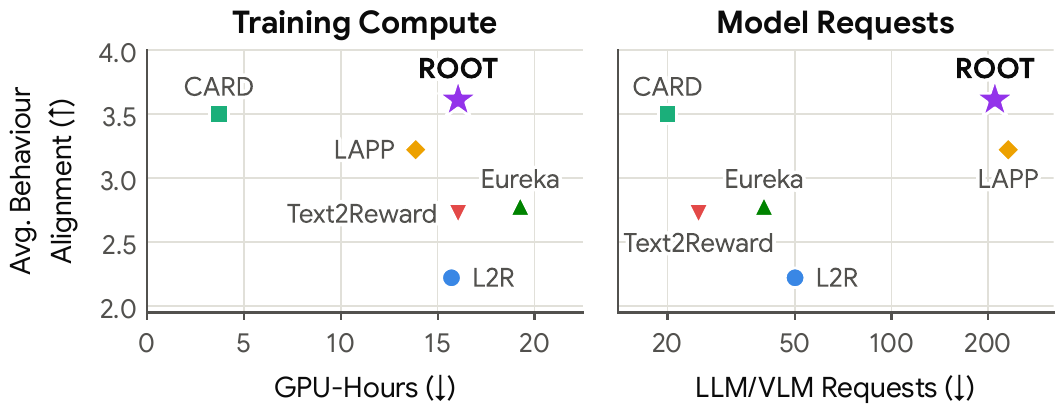}
        \par\smallskip
        {(b) Unitree Go2: Canter}
    \end{minipage}

    \caption{Avg. behavioral alignment versus training compute (GPU-hours, left) and LLM/VLM requests (right) on Ant crab-crawling and Go2 canter. Higher alignment and lower cost are better.}
    \label{fig:cost_analysis}
\end{figure*}

\subsection{RQ2: The Role of Tree Memory and Visual Feedback}
\label{sec:rq2}
We investigate how experimental memory (experience tree) and Vid-LLM visual feedback contribute to behavioral alignment and task performance. We compare full \textsc{ROOT} with two ablated variants: (i) \emph{without history}, which removes access to prior experiment records, and (ii) \emph{without history and visual feedback}, which additionally removes Vid-LLM feedback from reward search. We evaluate these configurations on Ant crab-crawling and spider-creeping, and Unitree Go2 cantering.

As shown in Table~\ref{tab:ablations}, full \textsc{ROOT} achieves the highest behavioral alignment and task accuracy on all three tasks. On an average, removing experimental history across tasks reduces the alignment score from 3.98 to 2.80 and task accuracy from 51.22\% to 10.74\%, corresponding to decreases of {1.18 score points} and {40.48 percentage points}, respectively. Additionally removing visual feedback lowers average alignment and accuracy to 1.84 and 1.91\%, respectively, yielding further decreases of {0.96 score points} and {8.83 percentage points}.

These results support the importance of retaining experimental history to inform subsequent reward refinements. Furthermore, visual feedback also provides useful guidance when history is unavailable, as removing it further degrades both metrics across all three tasks.

\begin{figure*}[t]
    \centering
    \vspace{0.5em}
    \includegraphics[width=\linewidth]{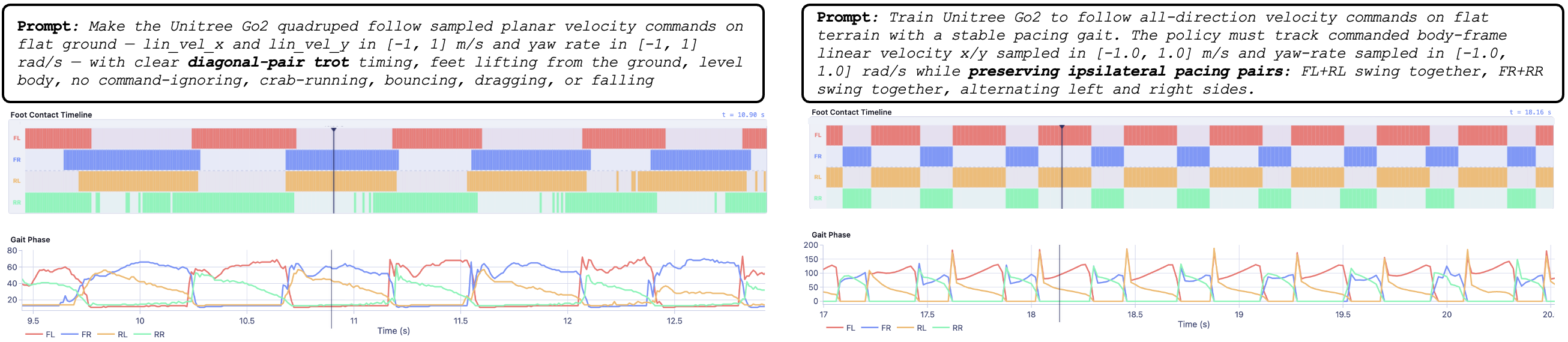}
    \caption{Foot contact timelines and foot-force signals for two Unitree Go2 locomotion gaits. (Left) \textbf{Trotting} on a real Unitree Go2 quadruped, showing characteristic alternating diagonal leg pairs (FL/RR and FR/RL) with regular, synchronised contact phases. (Right) \textbf{Pacing} in simulation, where ipsilateral leg pairs (FL/RL and FR/RR) move together, producing a distinctly different contact pattern.}
    \label{fig:trotting}
\end{figure*}

\begin{figure*}[t]
    \centering
    \vspace{0.5em}
    \includegraphics[width=\linewidth]{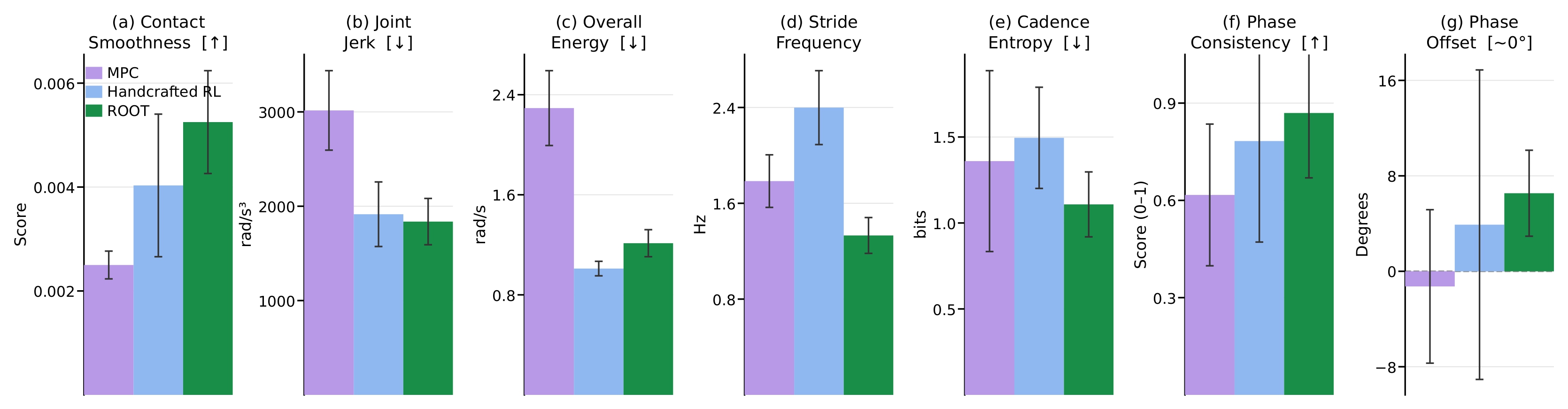}
    \caption{Gait quality metrics comparing three locomotion policies on a real Unitree Go2 robot across 7--8 trials per policy (mean $\pm$ SD). Metrics are grouped into three categories. \textbf{Motion Quality (a--c):} Contact Smoothness measures the regularity of foot touchdown transitions ($\uparrow$ better); Joint Jerk is the RMS rate of change of joint acceleration in rad/s$^3$ ($\downarrow$ better); Joint-motion magnitude is the RMS angular velocity across all joints in rad/s ($\downarrow$ better). \textbf{Gait Regularity (d--e):} Stride Frequency is the mean stepping cadence in Hz; Cadence Entropy measures irregularity in inter-stride intervals in bits ($\downarrow$ better). \textbf{Inter-leg Coordination (f--g):} Phase Consistency quantifies the stability of diagonal leg synchronization on a 0--1 scale ($\uparrow$ better); Phase Offset is the mean phase difference between diagonal leg pairs in degrees, where a trot ideally exhibits a phase offset of approximately $0^\circ$. These measurements characterize the recorded policies; joint angular velocity is a motion proxy and does not establish energy efficiency.
}
    \label{fig:trot_results}
\end{figure*}

\subsection{RQ3: Behavioral Alignment versus Search Cost}
\label{sec:rq3}
Fig.~\ref{fig:cost_analysis} compares behavioral alignment against policy-training GPU-hours and LLM/VLM requests per search run on Ant crab-crawling and Go2 canter.
\textsc{ROOT} achieves the highest mean Vid-LLM alignment on both tasks while using less training compute than Eureka, and less than LAPP on crab-crawling. Conversely,
CARD requires substantially less compute but achieves lower alignment. These results suggest that greater policy-training expenditure alone does not guarantee
better behavioral alignment.

\textsc{ROOT} allocates more model queries to observation and reasoning than the reward-code baselines, while using fewer queries than LAPP on both tasks. Taken together, the results show that \textsc{ROOT} achieves stronger measured alignment without incurring the highest training compute or query count. They support a favorable empirical balance between alignment and search cost, although they do not isolate the efficiency gains of individual components. Comparisons use method-specific budgets,
and query counts indicate interaction volume rather than monetary cost.

\subsection{RQ4: Sim-to-Real Transfer of Behavioral Alignment}
\label{sec:rq4}
We examine trotting on a physical Unitree Go2 and pacing in simulation to assess whether \textsc{ROOT} captures the coordination specified by each instruction. Fig.~\ref{fig:trotting} illustrates the resulting contact patterns. Trotting synchronizes diagonal FL/RR and FR/RL pairs (left), while pacing synchronizes same-side FL/RL and FR/RR pairs (right). These distinct pairings show that the learned behaviors reflect the requested gait structure beyond achieving forward motion.

We further transfer the selected trotting policy to a physical Go2 and compare it with Unitree's MPC controller and an RL controller trained using a hand-engineered locomotion reward. The hardware contact traces in Fig.~\ref{fig:trotting} (left) retain the alternating diagonal pattern, indicating that the learned coordination survives the transition from simulation to hardware. Furthermore, the quantitative results in Fig.~\ref{fig:trot_results} show improvements in both motion smoothness and coordination regularity. \textsc{ROOT}'s higher mean contact smoothness and lower joint jerk relative to MPC indicate gentler foot-contact transitions and less abrupt joint motion. Among the evaluated controllers, it achieves the lowest mean cadence entropy and highest mean phase consistency, indicating a more regular stepping rhythm with stable diagonal coordination. This regularity is maintained at a lower stride frequency than either baseline. MPC achieves a mean phase offset closer to zero, while \textsc{ROOT} maintains more consistent phase relationships over time. Together, these results show that \textsc{ROOT} preserves a repeatable trotting pattern, supporting transfer of \textsc{ROOT}'s ability for behavioral alignment on hardware.
\section{Conclusion}
\label{sec:conclusion}

We introduce \textsc{ROOT}, a framework for discovering reward functions that align learned policies with user-specified behaviors. \textsc{ROOT} combines visual feedback and multi-agent reasoning with a persistent experiment tree, using observed successes and failures to guide subsequent reward refinements. Experiments across multiple tasks and embodiments demonstrate improved behavioral alignment while maintaining competitive task performance and a favorable trade-off between alignment and search cost. Additionally, in our human survey, participants preferred behaviors produced by \textsc{ROOT} over those of the baselines when judging alignment with the specified task. Furthermore, deployment on a physical Unitree Go2 demonstrates that policies trained in simulation using rewards discovered by \textsc{ROOT} can transfer to the real world.
Future work could extend observation-guided reward discovery to manipulation and longer-horizon behaviors, where alignment requires coordinating multiple stages of a task. Another promising direction is to reuse behavioral insights stored in the experiment tree across related tasks and embodiments, allowing prior experience to guide reward discovery for new specifications. 
Overall, our findings highlight the value of grounding reward design in observed behavior and accumulated experimental evidence to bring learned policies into closer alignment with user intent.










\bibliographystyle{IEEEtran}
\bibliography{example}

\newpage
\clearpage
\useRomanappendicesfalse
\appendices

\section{Task Specifications}
\label{app:task-specs}

Table~\ref{tab:task_specs} lists the natural-language behavior specifications
for the seven simulated tasks. Each method receives the same task description
and environment context. The descriptions specify the intended behavior;
state-based evaluation criteria are defined separately in
Appendix~\ref{app:metrics}. FL, FR, RL, and RR denote the front-left,
front-right, rear-left, and rear-right legs, respectively.

\begin{table*}[t]
\centering
\caption{Natural-language behavior specifications and simulator backends.
The Go2 trot policy is also evaluated on a physical robot.}
\label{tab:task_specs}
\small
\setlength{\tabcolsep}{5pt}
\renewcommand{\arraystretch}{1.15}
\begin{tabularx}{\textwidth}{@{}p{0.16\textwidth}p{0.13\textwidth}X@{}}
\toprule
\textbf{Environment} & \textbf{Task} & \textbf{Behavior specification} \\
\midrule
Hopper (MJX) & Forward hopping &
Hop forward as fast as possible. \\
HalfCheetah (MJX) & Forward running &
Run forward as fast as possible. \\
Ant (MJX) & Spider-creep &
Make the ant creep forward like a spider, using all four legs in a coordinated
crawling gait while keeping its body close to the surface. \\
Ant (MJX) & Crab-crawl &
Make the ant crawl sideways like a crab, moving laterally using coordinated
leg movements while keeping its body low and stable. \\
Go2 (Isaac Lab) & Pace &
Train the Go2 to pace forward---legs on the same side (FL+RL, then FR+RR)
move together in a lateral two-beat gait with steady forward velocity. \\
Go2 (Isaac Lab) & Trot &
Train the Go2 to perform a trot gait---a two-beat diagonal pattern where
diagonally opposite legs (FL+RR, then FR+RL) move together in alternation
with steady forward velocity. \\
Go2 (MJX) & Canter &
Train the Go2 to perform a canter gait---a three-beat asymmetric running
pattern where one hind leg leads, followed by the other hind leg together
with the diagonal foreleg, then the remaining foreleg. \\
\bottomrule
\end{tabularx}
\end{table*}

The hardware experiment evaluates whether the simulation-trained policy
retains a stable trot after deployment. Its target behavior is straight-line
locomotion with alternating diagonal leg pairs. We treat hardware deployment
as a transfer evaluation of the learned behavior, rather than an additional
reward-search task.

\section{Implementation Details}
\label{app:baseline_details}

\subsection{Language Models and Reward Search}

All methods use Claude Sonnet 4.5 for language-model reasoning and reward-code
generation. The reward-generation temperature is $0.7$. In \textsc{ROOT}, the
Proposer, Critic, and Strategist use temperature $0.4$, while Gemini 3.1 Pro
provides rollout-video feedback during the Observe stage, with a maximum
output length of $1500$ tokens.

The reasoning agents use retrieval tools to inspect individual experiment
records and compare refinements across branches without loading the full
history into context. They also update an auxiliary insights document with
guidance distilled from these experiments. The selected parent's reward code
is supplied to the generator together with the refinement plan. Each candidate
policy is trained from scratch.

The search budget $B$ counts experiment attempts, including initialization
and unsuccessful execution attempts. Failed attempts retain their diagnostics
and remain available to subsequent reasoning. After search, \textsc{ROOT}
selects a successfully evaluated policy using the task-alignment descriptions
recorded in the tree. This search-time selection is separate from the final
VLM-panel and human evaluations described in Appendix~\ref{app:metrics}.

\subsection{Policy Training and Baselines}

Within each simulator backend, methods share the task descriptions,
environment configuration, PPO settings, and training steps per candidate.
MJX candidates receive $20$ million environment steps, except Go2 canter,
which receives $30$ million. Isaac Lab Go2 candidates receive approximately
$98$ million steps. Table~\ref{tab:ppo_hparams} summarizes the PPO settings
for the Ant and Go2 experiments.

\begin{table*}[t]
\centering
\caption{PPO settings for \textsc{ROOT}. Ant includes both crab-crawl and
spider-creep. MJX tasks use Brax PPO; Isaac Lab tasks use RSL-RL PPO.
Training budgets are environment transitions per reward candidate.
Dashes indicate unspecified or inapplicable settings.}
\label{tab:ppo_hparams}
\small
\setlength{\tabcolsep}{3.5pt}
\renewcommand{\arraystretch}{1.12}
\begin{tabular*}{\textwidth}{@{\extracolsep{\fill}}lcccc@{}}
\toprule
\textbf{Parameter} &
\shortstack{\textbf{Ant / Hopper}\\\textbf{(MJX)}} &
\shortstack{\textbf{HalfCheetah}\\\textbf{(MJX)}} &
\shortstack{\textbf{Go2 canter}\\\textbf{(MJX)}} &
\shortstack{\textbf{Go2 pace/trot}\\\textbf{(Isaac Lab)}} \\
\midrule
Parallel environments & $4096$ & $4096$ & $8192$ & $4096$ \\
Episode limit (control steps) & $1000$ & $1000$ & $1000$ & $1000$ \\
Transitions per candidate &
$2\times10^7$ & $10^7$ & $3\times10^7$ & $98{,}304{,}000$ \\
\addlinespace[3pt]
Training iterations (RSL-RL) & --- & --- & --- & $1000$ \\
Learning rate &
$3\times10^{-4}$ & $3\times10^{-4}$ & $3\times10^{-4}$ & $10^{-3}$ \\
Learning-rate schedule & Constant & Constant & Constant & Adaptive \\
Discount $\gamma$ & $0.97$ & $0.97$ & $0.97$ & $0.99$ \\
GAE $\lambda$ & $0.95$ & $0.95$ & $0.95$ & $0.95$ \\
PPO clipping parameter & $0.2$ & $0.2$ & $0.2$ & $0.2$ \\
\addlinespace[3pt]
Sequences per minibatch (Brax) & $1024$ & $1024$ & $256$ & --- \\
Transitions per minibatch &
$10{,}240$ & $10{,}240$ & $5120$ & $24{,}576$ \\
Minibatches per epoch & $32$ & $32$ & $32$ & $4$ \\
Optimization epochs per batch & $8$ & $8$ & $4$ & $5$ \\
Rollout sequence length & $10$ & $10$ & $20$ & $24$ \\
Transitions per PPO batch &
$327{,}680$ & $327{,}680$ & $163{,}840$ & $98{,}304$ \\
\addlinespace[3pt]
Reward scaling & $0.1$ & $0.1$ & $1.0$ & --- \\
Entropy coefficient & $0$ & $0$ & $10^{-2}$ & $10^{-2}$ \\
Observation normalization & Yes & Yes & Yes & --- \\
\bottomrule
\end{tabular*}
\par\vspace{3pt}
\begin{minipage}{\textwidth}
\footnotesize
Brax's \texttt{batch\_size} counts rollout sequences per minibatch;
each sequence contains the listed rollout length. Thus, transitions per
minibatch equal \texttt{batch\_size} times rollout length, and transitions
per PPO batch additionally multiply by the number of minibatches.
For RSL-RL, the corresponding counts are
$4096\times24/4=24{,}576$ transitions per minibatch and
$4096\times24=98{,}304$ transitions per iteration.
\end{minipage}
\end{table*}

We compare with Eureka, Text2Reward, L2R, CARD, and LAPP. Candidate counts
and policy-selection rules follow each method's protocol, rather than a
common candidate cap. The LAPP adaptation uses only the preference reward,
$r=\beta r_p$, with the handcrafted environment reward removed. Its total
environment-step budget is matched to that of \textsc{ROOT}. All selected
policies are assessed with the same final evaluation protocol within each task.

\subsection{Ablation Configurations}

The ablations in RQ2 evaluate Ant crab-crawl, Ant spider-creep, and Go2
canter. The \emph{No History} variant removes access to prior experiment
records and accumulated insights while retaining visual feedback on the
current rollout. The \emph{No Vid-LLM and History} variant additionally
removes rollout-video feedback, leaving reward search without either source
of behavioral evidence. Both variants use the same task definitions and final
evaluation metrics as the full method.

This comparison measures the effect of history when visual feedback is
available, followed by the effect of visual feedback when history is absent.
It does not separately measure the effect of removing visual feedback from
the full history-enabled method.

\section{Evaluation Protocols}
\label{app:metrics}

We evaluate task performance using simulator states, behavioral alignment
using an external VLM panel, and perceived adherence using human preferences.
These measurements capture complementary aspects of behavior. In simulation,
each selected policy is evaluated over a $1000$-step horizon with three seeds.
The mean and standard deviation of task performance describe rollout
variability for that policy, rather than variability across independently
repeated reward searches.

\subsection{State-Based Task Performance}
\label{app:metrics:kinematic}

Task metrics are computed independently of the generated reward. For Hopper
and HalfCheetah, the native score is mean forward velocity multiplied by mean
survival. For the other tasks, locomotion completeness accuracy measures
task-specific coordination during locomotion. Because its definition differs
across embodiments and gaits, we compare accuracy within each task.

\paragraph{Contact extraction}
For contact-based MJX evaluation, a foot is marked as in contact
when its height is below $0.05$\,m. If this threshold detects contact for
fewer than half of the feet, an adaptive per-foot threshold uses the lowest
$30\%$ of that foot's height range. Contact signals are smoothed with a
$5$-step moving window. Feet are identified from body names, with initial
position quadrants used as a fallback. Locomotion is considered active when
horizontal speed exceeds $0.1$\,m/s.

\paragraph{Ant crab-crawl and spider-creep}
The Ant score measures agreement with the target contact-coordination pattern
during active locomotion. Spider-creep requires forward motion
($v_x>0.1$\,m/s) with diagonal coordination (FL/RR and FR/RL).
Crab-crawl requires lateral motion ($|v_y|>0.1$\,m/s) with ipsilateral
coordination (FL/RL and FR/RR), together with checks on vertical and angular
motion. These contact criteria provide a measurable proxy for coordination;
the VLM and human evaluations additionally assess posture and movement style.

\paragraph{Go2 pace and trot}
Accuracy combines survival, velocity tracking, contact coordination, and
smoothness:
\begin{equation}
  A_{\mathrm{pace/trot}}
  =100\,S\,V\,C\,M,
\end{equation}
where $S$, $V$, $C$, and $M$ denote the component scores used by the
evaluator. Pace targets synchronized ipsilateral pairs, whereas trot targets
synchronized diagonal pairs, with alternating support between the two pairs.
Evaluation uses a forward command of $0.5$\,m/s with external pushes disabled.

\paragraph{Go2 canter}
Accuracy is the fraction of detected gait cycles that match the target
three-beat footfall pattern, weighted by survival:
\begin{equation}
  A_{\mathrm{canter}}
  =100\,S\,\frac{N_{\mathrm{match}}}{N_{\mathrm{detected}}}.
\end{equation}
Here, $N_{\mathrm{detected}}$ is the number of detected cycles and
$N_{\mathrm{match}}$ counts those satisfying the evaluator's canter criterion.

\subsection{VLM Behavioral Alignment}
\label{app:vlm_eval}

Final alignment is assessed by Seed 2.1 Turbo, Gemma 4 31B, and GLM-4.6V,
a panel separate from the Gemini 3.1 Pro model used for search-time
observation. Each judge receives the task description, three-view rollout
videos, and simulator-derived facts. The views show the robot from the left,
right, and above. Method identities, reward code, and training curves are not
shown to the judges.

Judges score posture and style and limb coordination on integer scales from
$1$ to $5$, with a short justification. The reported alignment score averages
the two dimensions across judges and evaluation rollouts. VLM ratings are
interpreted alongside state-based metrics because visually plausible movement
does not necessarily imply the correct contact pattern.

\subsection{Human Preferences}
\label{app:human_eval}

Human evaluation focuses on Ant crab-crawl, Ant spider-creep, and Go2 canter.
Raters receive the behavior description and a pair of rollout videos showing
\textsc{ROOT} and a baseline. Videos are labeled A and B, with randomized
ordering and no method identifiers. The primary question asks which video
better matches the requested behavior. Raters may select either video or
indicate no clear distinction.

For a task, the preference for \textsc{ROOT} against pooled baselines is
\begin{equation}
  P_{\textsc{ROOT}}
  =\frac{W}{W+L},
\end{equation}
where $W$ and $L$ are the numbers of non-tied judgments preferring
\textsc{ROOT} and a baseline, respectively. Ties are excluded from this
percentage. Pooling uses the judgments across baseline comparisons for the
same task.

\subsection{Reward-Search Cost}
\label{app:search_cost}

RQ3 compares alignment with policy-training GPU-hours and the number of
LLM/VLM requests per search run on Ant crab-crawl and Go2 canter. GPU-hours
measure policy-training compute, while request counts measure model
interaction volume. Candidate budgets and selection procedures remain
method-specific. These comparisons characterize the observed balance between
alignment and search cost; they do not isolate the cost contribution of an
individual \textsc{ROOT} component or equate model requests with monetary cost.

\section{Physical-Robot Experiments}
\label{app:physical_exp}

\subsection{Policy Training and Deployment}

We deploy the learned trot policy on a Unitree Go2 with $12$ actuated joints,
three per leg. Training uses PPO in Isaac Lab with $4096$ parallel environments
and an asymmetric actor--critic. The critic receives privileged observations,
including ground-truth base linear velocity. The actor uses a $45$-dimensional
observation comprising joint positions and velocities, base angular velocity,
projected gravity, commanded velocities, and the previous action.

Training randomizes friction coefficients uniformly over $[0.3,1.2]$ and
base-mass perturbations over $[-1,+3]$\,kg. A PD actuator model includes a
torque--speed limit and action-execution delay. After training, the actor is
exported to ONNX with a configuration specifying observation scales and
command ranges.

The policy is deployed without fine-tuning. Inference runs on a desktop
computer at $50$\,Hz through the Unitree SDK low-level interface. Joint
positions and velocities come from encoders, base angular velocity from the
IMU, and projected gravity from its orientation estimate. Deployment uses
the exported observation and action scales. Commands are restricted to
straight-line locomotion, with forward velocity in $[-0.6,0.6]$\,m/s and
zero lateral and yaw velocity.

\subsection{Controller Comparison}

We compare the transferred policy with Unitree's MPC controller and an RL
controller trained with a hand-engineered locomotion reward. Each trial begins
from standing and ends when the robot leaves the camera view or requires
operator intervention. The current comparison contains $7$--$8$ trials per
controller and reports mean and standard deviation across trials. These
hardware statistics are separate from the three-seed simulation evaluation.

\subsection{Gait-Quality Metrics}

The seven metrics reported in the main paper characterize contact transitions,
joint motion, stepping rhythm, and diagonal coordination. They are computed
from recorded foot-force and joint-angle signals.

\paragraph{Contact smoothness}
This score summarizes the rate of change of the foot-force signal:
\begin{equation}
  Q_{\mathrm{contact}}=\frac{1}{1+\bar{s}_F},
\end{equation}
where $\bar{s}_F$ is the mean force-signal slope used by the evaluator.
Higher scores indicate less abrupt measured contact transitions. The score
depends on force scaling and temporal sampling, which must be consistent
across controllers.

\paragraph{Joint jerk}
For joint angle $q_j(t)$, jerk is the third time derivative,
$\dddot{q}_j(t)$. We report root-mean-square jerk across joints and samples
in $\mathrm{rad/s^3}$. Lower values indicate less abrupt joint motion.

\paragraph{Joint-motion magnitude}
We report the root-mean-square joint angular velocity in $\mathrm{rad/s}$.
This measures the magnitude of joint motion. Without torque or power
measurements, it does not establish mechanical work or energy efficiency.

\paragraph{Stride frequency}
Touchdown events are detected from threshold crossings in foot-force signals.
Stride frequency summarizes the stepping cadence in Hz. A higher or lower
frequency is descriptive and is not, by itself, evidence of a better gait.

\paragraph{Cadence entropy}
The entropy of the inter-stride interval distribution is
\begin{equation}
  H=-\sum_b p_b\log_2 p_b,
\end{equation}
where $p_b$ is the fraction of intervals in bin $b$, and zero-probability
terms contribute zero. Lower entropy indicates more concentrated stride
timing under a common discretization.

\paragraph{Phase consistency}
This $0$--$1$ score measures how consistently diagonal leg pairs maintain
their relative phase across short recording windows. Higher values indicate
less temporal drift in diagonal coordination.

\paragraph{Diagonal phase offset}
Cross-correlation of foot-force signals estimates the timing lag within
FL/RR and FR/RL pairs. The lag is expressed as a phase difference in degrees
relative to stride timing. Values near $0^\circ$ indicate synchronized
diagonal legs. Phase offset describes average synchronization, whereas phase
consistency describes its stability over time.

\section{Reasoning and Evaluation Prompts}
\label{app:prompts}

Figures~\ref{fig:root_proposer_prompt}--\ref{fig:root_evaluator_prompt}
present condensed prompt templates for reasoning, reward generation, and
final behavioral evaluation. Bracketed text denotes task-specific inputs.
The reasoning agents inspect the experience tree, individual experiment
records, and accumulated insights through retrieval tools. The templates
below summarize the MJX reasoning roles; Isaac Lab uses locomotion-specific
versions and names the Proposer the \emph{Scientist}. Its Strategist also
selects a reward-design archetype. The final behavioral evaluator is
distinct from the Vid-LLM that provides feedback during reward search.

Search-time observation and final behavioral evaluation serve different
purposes. The Observe stage supplies feedback for subsequent reward
refinement, whereas the external panel scores the selected policies after
search. The annotator combines search-time observations, diagnostics,
refinement plans, parent findings, and current insights into the findings
stored with each experiment.

\begingroup
\setlength{\parskip}{0pt}
\captionsetup{position=bottom,skip=5pt}


\begin{ROOTpromptbox}{ROOTReason}{Reason: Proposer}
\ROOTpromptfield{System Prompt}

You are a reward-function research proposer analyzing an experiment tree.
Identify cross-branch patterns, partial successes, and a promising next
experiment.

Inspect the full tree using \texttt{get\_thought\_tree()} and retrieve
relevant records using \texttt{get\_node\_detail()}. Compare task-alignment
observations with survival, velocity, posture, and training diagnostics.
When visual descriptions conflict with measured survival or motion, use
the measurements to resolve the discrepancy. A surviving policy with low
forward velocity is alive but not moving; a policy that terminates after
only a few steps is failing even if the video summary suggests otherwise.

Distinguish useful motor coordination without translation from true
inactivity. High limb range of motion and periodicity indicate a valuable
partial success even when forward velocity is near zero. If action
magnitudes are large but the visual summary reports only twitching, inspect
the node in detail. Compare reward-component training curves to identify
convergence, plateaus, and terms that dominate optimization.

Read accumulated insights and avoid repeating falsified hypotheses. A
branch with at least two children showing only lateral changes or
regressions is stuck: choose another parent or a fundamentally different
reward structure. Returning to the root is allowed when a clean change of
direction is needed.

Conclude with \emph{Task Alignment Status}, \emph{Stuck Branches},
\emph{Cross-Branch Patterns}, \emph{Partial Successes Worth Exploring}, and
\emph{Proposed Direction}. Specify the parent, changes, testable hypothesis,
and evidence that the proposal does not repeat a failed pattern.

\ROOTpromptfield{User Prompt}

Analyze the experiment journal for the following task and propose the next
experiment. Begin by inspecting the full tree.

\textit{[Task description or grounded behavioral specification]}
\end{ROOTpromptbox}
\par\nopagebreak[4]
\captionof{figure}{Condensed Proposer prompt for MJX reward search.}
\label{fig:root_proposer_prompt}


\begin{ROOTpromptbox}{ROOTReason}{Reason: Critic}
\ROOTpromptfield{System Prompt}

You are a critical reviewer challenging the Proposer's analysis of a
reward-search tree for \textit{[Task description]}. Check whether the
proposal addresses the requested behavior rather than merely increasing
native return.

Verify claims against quantitative diagnostics and experiment records.
Challenge unsupported interpretations of training dynamics, repeated
hypotheses, and proposals that continue a stuck branch. Retrieve specific
nodes to support your objections.

You must identify at least one overlooked node and at least one flaw or
weak assumption in the proposed direction. Return your task-alignment
assessment, overlooked nodes, challenges to the hypothesis, relevant prior
experiments, and concerns that the Strategist must address.

\ROOTpromptfield{User Prompt}

Review the Proposer's latest position using the debate history and
retrieval tools. If earlier concerns have been resolved, acknowledge this
and identify new concerns or deeper issues.

\textit{[Proposer's analysis and debate history]}
\end{ROOTpromptbox}
\par\nopagebreak[4]
\captionof{figure}{Condensed Critic prompt for MJX reward search.}
\label{fig:root_critic_prompt}


\begin{ROOTpromptbox}{ROOTReason}{Reason: Strategist}
\ROOTpromptfield{System Prompt}

You are a strategist synthesizing a Proposer--Critic debate into one
concrete reward-refinement experiment. Your prescription must be
self-contained because the generator does not receive the debate.

Address each remaining Critic concern and justify the selected parent.
Include relevant parent-code snippets, the components to preserve or
change, exact formulas and coefficient ranges, observation indices or
information keys, and the expected magnitude of each term. Ground the
changes in the recorded evidence.

Return the following sections: \emph{Selected Parent}, \emph{Hypothesis},
\emph{Prescription for GENERATE}, \emph{Addressing Critic's Concerns}, and
\emph{Current Best Node}. State what outcomes would confirm or refute the
hypothesis. The current best node may differ from the selected parent.
Use \texttt{update\_insights()} when the evidence warrants an update to the
accumulated principles.

\ROOTpromptfield{User Prompt}

Synthesize the debate into a complete prescription for the next experiment.
Use retrieval tools to review the cited nodes and explicitly address the
Critic's remaining concerns. Optimizing native return alone is insufficient.

\textit{[Task description; complete debate transcript]}
\end{ROOTpromptbox}
\par\nopagebreak[4]
\captionof{figure}{Condensed Strategist prompt for MJX reward search.}
\label{fig:root_strategist_prompt}


\begin{ROOTpromptbox}{ROOTExecute}{Execute: Reward Generator}
\ROOTpromptfield{System Prompt: MJX}

You are an expert reinforcement-learning reward engineer. Write a Python
reward function for a fresh PPO policy using only the available observation,
action, and information signals. Use the provided indices and prefer
information keys when available. Return a finite scalar reward and a
dictionary of named component values. Use only the supplied NumPy and
\texttt{math} interfaces; include no imports.

\ROOTpromptfield{User Prompt: MJX}

Write a reward function that follows the Strategist's prescription exactly.
Preserve components designated to remain unchanged and respect the
specified formulas and coefficient ranges.

\textit{[Task specification; environment specification; Strategist's
prescription, including parent reference code]}

Return only JSON with \texttt{reasoning}, \texttt{component\_formulas}, and
\texttt{code}. The code implements
\texttt{compute\_reward(obs, action, raw\_reward,}\par
\texttt{terminated, truncated, info)} and returns
\texttt{(total\_reward, components\_dict)}.

For initialization, replace the prescription with a bootstrap strategy and
available prior insights; no parent experiment is provided.

\ROOTpromptfield{System and User Prompts: Isaac Lab}

Reason about locomotion failure modes and design reward components that
address them. Follow the supplied Go2 environment-data schema, gait-specific
constraints, and design archetype. When supplied, the Strategist's
prescription takes priority over the archetype.

\textit{[Task description; Go2 environment-data schema and constraints;
archetype; optional prescription and prior video feedback]}

\tcbbreak

Return a JSON code block containing \texttt{components}. Each component
specifies \texttt{name}, \texttt{default\_weight}, and \texttt{code}; custom
code returns a tensor of shape \texttt{[num\_envs]}. Use the designated
component names for contact-based terms and obey the required component
count of 6--12.
Include the mandatory orientation, base-height, feet-air-time,
gait-contact-phase, and lateral-foot-spacing components.
\end{ROOTpromptbox}
\par\nopagebreak[4]
\captionof{figure}{Condensed generation templates and output interfaces for
MJX and Isaac Lab. In MJX, parent code is included in the prescription rather
than passed as a separate experiment record.}
\label{fig:root_generator_prompt}


\begin{ROOTpromptbox}{ROOTObserve}{Evaluate: Behavioral Evaluator}
\ROOTpromptfield{System Prompt}

You are an expert robotics evaluator scoring locomotion policies.

\ROOTpromptfield{User Prompt}

Evaluate the rollout for \textit{[Task description]} using all supplied
camera views, or temporally ordered frames. Judge a reasonable approximation
of the requested behavior given the robot's physical constraints.

The camera follows the robot: assess travel relative to the floor rather
than screen position. Ignore the frozen initial padding. Use the supplied
simulator facts about termination and torso flips to resolve apparent
falls; your fall description must agree with these facts.

\textit{[Rollout videos or ordered frames; simulator episode facts]}

First describe \texttt{falls}, \texttt{progress}, \texttt{torso}, and
\texttt{legs}. Then assign integer scores from 1 to 5:

\textbf{Posture and movement style.}
1: fallen, flipped, stuck, or no meaningful motion;
2: upright but wrong direction or orientation;
3: approximately correct motion with noticeable posture issues;
4: good posture and direction with minor deviations;
5: close match to the requested posture and movement style.

\textbf{Limb coordination.}
1: chaotic, flailing, or still;
2: leg motion without a recognizable coordination pattern;
3: rhythmic motion without a clear match to the target gait;
4: coordinated motion mostly resembling the target gait;
5: clear, sustained coordination closely matching the target gait.

Return only JSON with \texttt{observations}, \texttt{posture\_style},
\texttt{limb\_coordination}, and \texttt{reasoning}. Keep the explanation to
one or two sentences linking the observations to the scores.
\end{ROOTpromptbox}
\par\nopagebreak[4]
\captionof{figure}{Condensed final behavioral-evaluation template
(\texttt{v3}), including observation-first scoring and simulator episode facts.}
\label{fig:root_evaluator_prompt}
\label{app:vlm_prompt}

\endgroup

\end{document}